\documentclass[10pt,twocolumn,letterpaper]{article}

\usepackage{cvpr}              

\usepackage{graphicx}
\usepackage{amsmath}
\usepackage{amssymb}
\usepackage{booktabs}
\usepackage[accsupp]{axessibility}

\usepackage{xcolor}         
\usepackage{cite}
\usepackage{times}
\usepackage{epsfig}
\usepackage{algorithm,algpseudocode}
\usepackage{multirow}
\usepackage{mathtools, nccmath}
\usepackage{color,soul}
\usepackage{subcaption}
 \usepackage{amsmath}
 
\DeclarePairedDelimiter\abs{\lvert}{\rvert}%

\DeclareUnicodeCharacter{2212}{-}

\usepackage{subcaption}
\usepackage{makecell}
\usepackage{multirow}
\usepackage{ dsfont }
\definecolor{darkBlue}{rgb}{0.5,0.6,1}
\definecolor{LightCyan}{rgb}{0.5,0.8,1}
\definecolor{yellow}{rgb}{1,1,0.6}

\definecolor{green}{rgb}{0,0.4,0.2}
\definecolor{red}{rgb}{0.8,0,0.1}

\newcommand{\beginsupplement}{%
        \setcounter{table}{0}
        \setcounter{figure}{0}
        \setcounter{section}{0}
}

\usepackage[pagebackref,breaklinks,colorlinks]{hyperref}

\usepackage[capitalize]{cleveref}
\crefname{section}{Sec.}{Secs.}
\Crefname{section}{Section}{Sections}
\Crefname{table}{Table}{Tables}
\crefname{table}{Tab.}{Tabs.}

\def\cvprPaperID{3265} 
\def\confName{CVPR}
\def\confYear{2022}

\begin{document}

\title{Fire Together Wire Together: \\A Dynamic Pruning Approach with Self-Supervised Mask Prediction}

\author{Sara Elkerdawy$^1$ \and Mostafa Elhoushi$^2$ \and Hong Zhang$^1$ \and Nilanjan Ray$^1$\\
$^1$University of Alberta, $^2$Toronto Heterogeneous Compilers Lab, Huawei\\
{\tt\small \{elkerdaw, hzhang, nray1\}@ualberta.ca}
}
\maketitle

\begin{abstract}
   Dynamic model pruning is a recent direction that allows for the inference of a different sub-network for each input sample during deployment. However, current dynamic methods rely on learning a continuous channel gating through regularization by inducing sparsity loss. This formulation introduces complexity in balancing different losses (e.g task loss, regularization loss). In addition, regularization based methods lack transparent tradeoff hyperparameter selection to realize a computational budget. Our contribution is two-fold: 1) decoupled task and pruning losses. 2) Simple hyperparameter selection that enables FLOPs reduction estimation before training. Inspired by the Hebbian theory in Neuroscience: “neurons that fire together wire together”, we propose to predict a mask to process $k$ filters in a layer based on the activation of its previous layer. We pose the problem as a self-supervised binary classification problem. Each mask predictor module is trained to predict if the log-likelihood for each filter in the current layer belongs to the top-$k$ activated filters. The value $k$ is dynamically estimated for each input based on a novel criterion using the mass of heatmaps. We show experiments on several neural architectures, such as VGG, ResNet and MobileNet on CIFAR and ImageNet datasets. On CIFAR, we reach similar accuracy to SOTA methods with 15\% and 24\% higher FLOPs reduction. Similarly in ImageNet, we achieve lower drop in accuracy with up to 13\% improvement in FLOPs reduction. Code is available at  \href{https://github.com/selkerdawy/FTWT}{https://github.com/selkerdawy/FTWT}

\end{abstract}


\begin{figure}%
\centering

\includegraphics[width=\columnwidth]{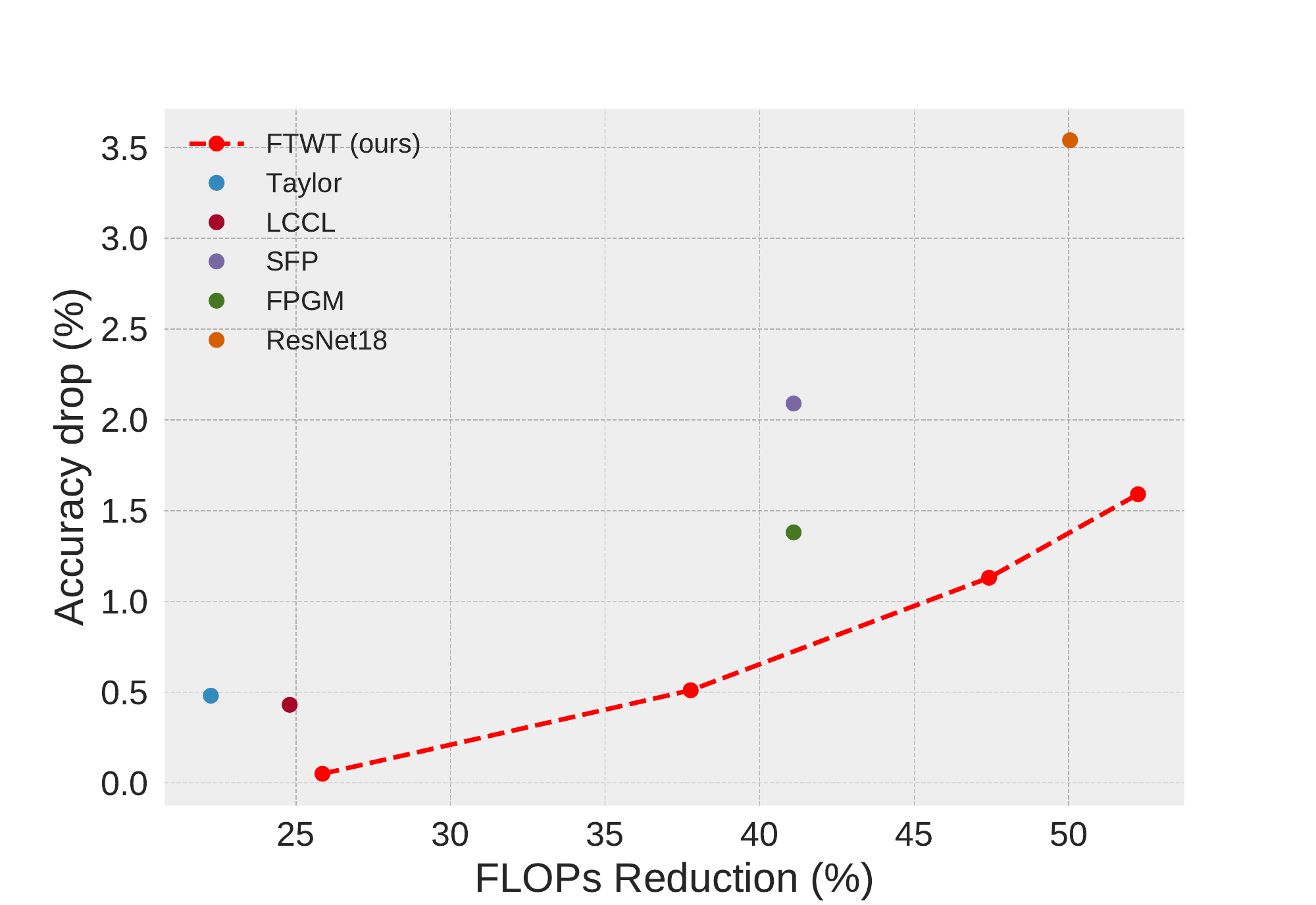}%
\caption{FLOPs reduction vs accuracy drop from baselines for various dynamic and static models on ResNet34 ImageNet.}
\label{fig:resnet34}
\end{figure}

\begin{figure*}%
\centering
\begin{subfigure}[b]{0.9\columnwidth}
\includegraphics[width=\columnwidth]{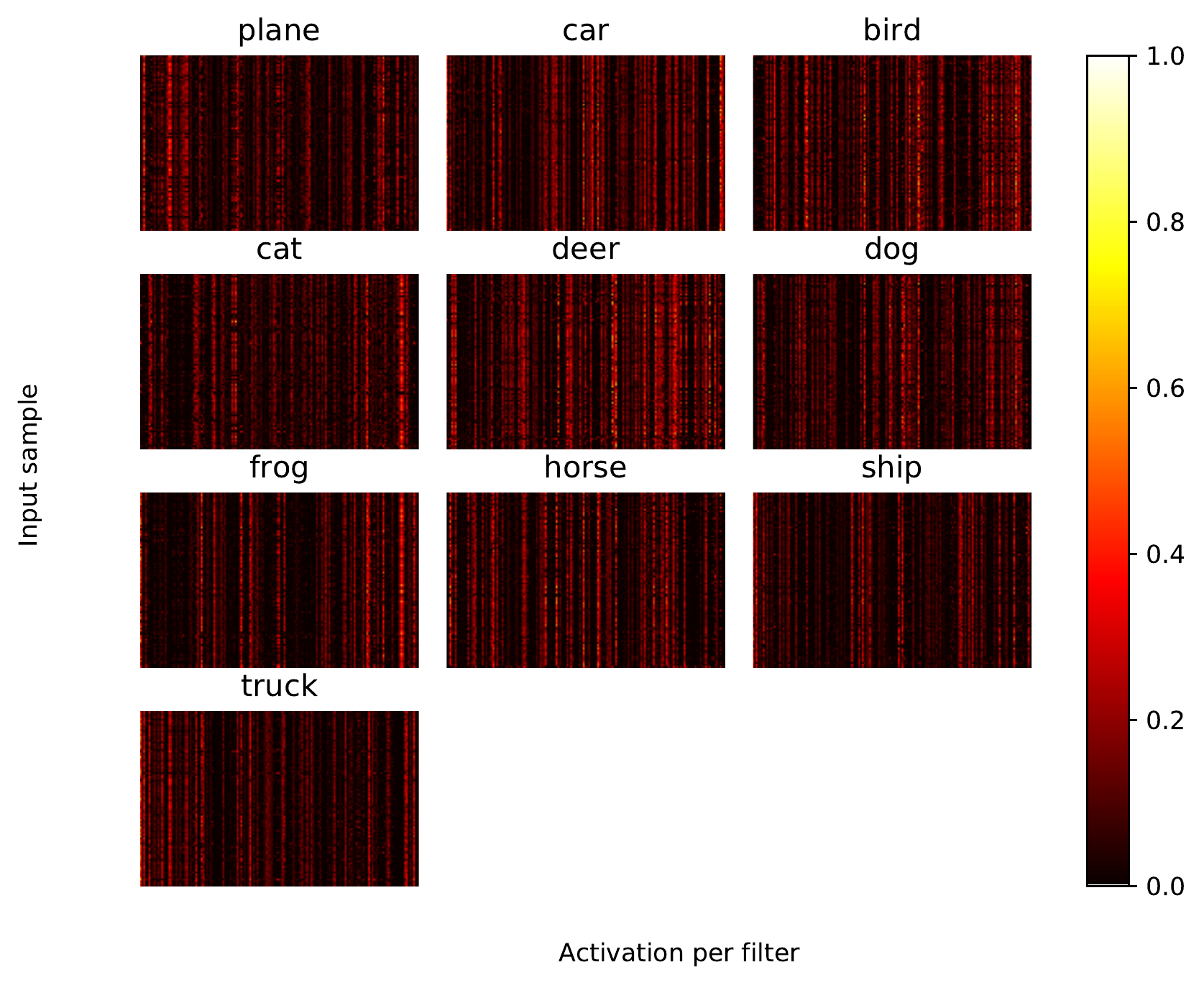}%
\caption{Last convolutional layer}%
\label{fig:clustera}%
\end{subfigure}
\begin{subfigure}[b]{0.9\columnwidth}
\includegraphics[width=\columnwidth]{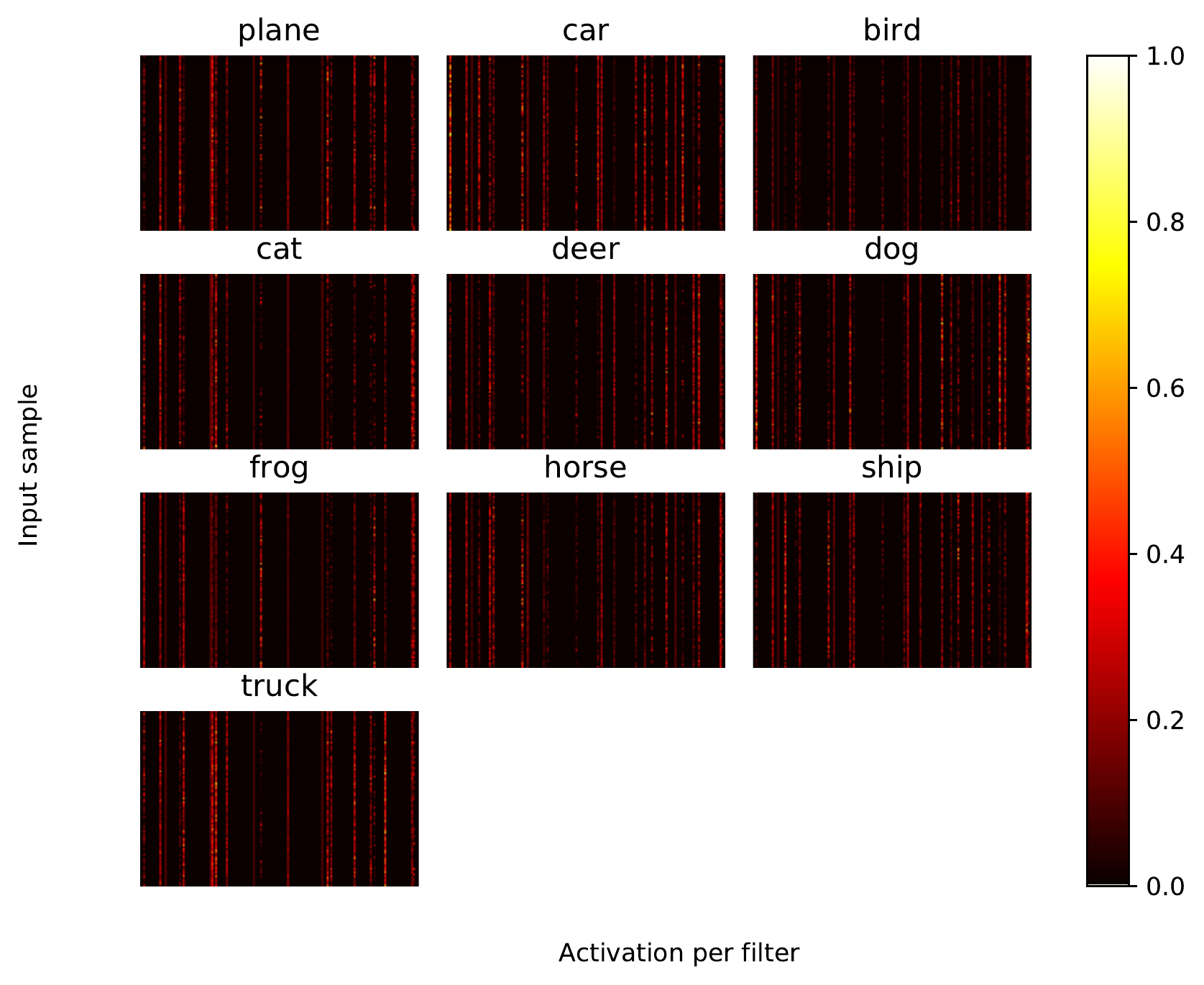}%
\caption{8th depthwise convolutional layer}%
\label{fig:clusterb}%
\end{subfigure}\hfill%

\caption{Maximum activations in all features at the last convolutional layer and a middle layer in mobilenetv1 CIFAR-10. Each row in a subplot represents an input sample. Samples that belong to the same class activate same group of filters. Better visualized in color.}
\label{fig:cluster}
\end{figure*}

\section{Introduction}

Convolutional Neural Networks (CNNs) showed unprecedented growth over the past decade which represented the state-of-the-art in many fields. However, CNNs require substantially large computation and memory consumption which limits deployment on edge and embedded platforms. There are many advances in model compression research including manually designed lightweight models \cite{howard2017mobilenets,hu2018squeeze}, low-bit precision \cite{li2019additive,zhang2018lq}, architecture search \cite{tan2019mnasnet,cai2019once}, and model pruning \cite{taylor,liu2018rethinking,frankle2018lottery,li2016pruning}. Most of the compression techniques are agnostic to the input data and optimize for a \textit{static} efficient model. Recent efforts in pruning literature propose to keep the backbone for the baseline model as a whole and do inference using different sub-networks conditioned on the input. This is known as dynamic pruning, where different routes are activated based on the input which allows higher degree of freedom and more flexibility in comparison to static pruning.

Current dynamic pruning approaches typically introduce a regularization term to induce sparsity over a continuous parameter for channel gating/masking \cite{FBS,dong2017more,wang2020dynamic}. Others adopt policy gradient introduced in reinforcement learning \cite{williams1992simple} to learn different routes. These methods require careful tuning in training to tackle issues such as training stability with schedule annealing \cite{wang2020dynamic}, biased training handling \cite{hua2018channel}, or predefined pruning ratio per layer \cite{FBS,lin2017runtime}. Also, as noted in \cite{dong2017more}, additional sparsity loss degrades task loss as it is difficult to balance the task loss and the pruning loss especially under high pruning ratio as shown in Figure \ref{fig:resnet34}. Moreover, the FLOPs reduction of these dynamic methods is dependent on the target sparsity preset hyperparameter. This hyperparameter selection lacks transparent relation between sparsity hyperparameter and the reached FLOPs; thus, hinders practical efficient training with many iterations of trial and error to achieve a target FLOPs reduction.

In this paper, we tackle these issues by formulating the problem as a self-supervised binary classification task. We generate the binary mask of the current layer (wiring) based on the activation (firing) of the previous layer. We draw inspiration from the Hebbian theory \cite{lowel1992selection} in Neuroscience with a twist that we enforce this wiring-firing relation instead of a study of causation as in the theory. Figure \ref{fig:cluster} plots the maximum response for each filter (x-axis) of the last convolutional layer and a middle layer of MobileNet-V1 for random input samples (y-axis) grouped by their class. The plot shows that samples that belong to the same class tend to activate the same combination of filters and thus we only need to process a handful of the filters. It is worth noting that the number of clusters vary per layer. Similar to other dynamic pruning methods, we learn a decision head for channel gating. However, we learn the gating using binary cross entropy loss per channel. Each layer predicts the filters which are most likely to be highly activated given the layer's input activations. We generate ground truth binary masks per layer based on the mass of the heatmap per sample. This formulation provides advantages in two aspects. First, the channel gating loss implicitly complies and adapts to the backbone's status which stabilizes training in comparison to the case with sparsity regularization or RL-based training. Second, reduction in FLOPs can be estimated before training, as the target mask is controlled by the generated ground truth mask which gives an estimate on the reduction. This simplifies the hyperparameter selection that controls pruning ratio. The main contributions are summarized as follows:

\begin{itemize}
  \item A novel loss formulation with self-supervised ground truth mask generation that is stochastic gradient descent (SGD) friendly with no gradient weighting tricks. 
  \item We propose a novel dynamic signature based on the heatmap mass without a pre-defined pruning ratio per layer.
  \item Simple hyperparameter selection that enables FLOPs reduction estimation before training. This simplifies realizing a prior budget target with bounded hyperparameter search space.
\end{itemize}

\begin{figure*}[!htbp]
\begin{center}

\includegraphics[width=0.90\linewidth]{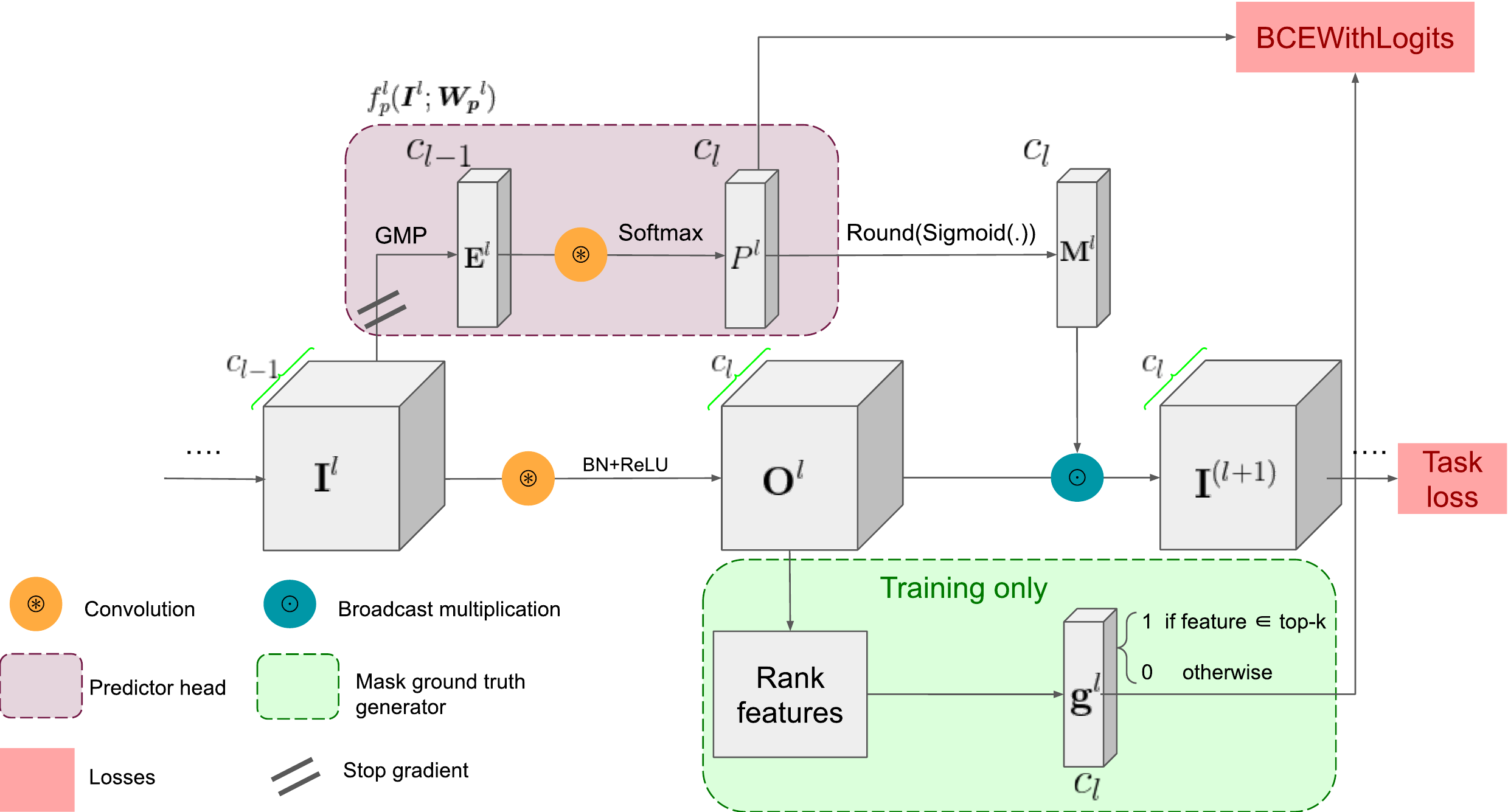}
\end{center}
\caption{Proposed pipeline for training dynamic routing for one layer. For a layer $l$, prediction head $f_{p}^l(\boldsymbol{I}^l;\boldsymbol{W_p}^l)$ takes an input $\boldsymbol{I}^l$, applies global max pooling (GMP), normalizes with Softmax, then feeds to 1x1 convolution to generate logits $\boldsymbol{P^l}$ for the binary mask $\boldsymbol{M}^l$. Binary Cross Entropy (BCEWithLogits) loss penalizes the mask prediction based on the top-$k$ obtained from the unpruned feature maps $\boldsymbol{O}^l$.}

\label{fig:dynamic}
\end{figure*}

\section{Related Work}

\textbf{Static Pruning.} Static pruning removes weights in the offline training stage and apply the same compressed model to all samples. Unstructured pruning \cite{frankle2018lottery,han2015learning,louizos2017learning,sharify2019laconic} targets removing individual weights with minimal contribution. The limitation of the unstructured weight pruning is that dedicated hardware and libraries \cite{sharify2019laconic} are needed to achieve speedup from the compression. Structured pruning is becoming a more practical solution where filters or blocks are ranked and pruned based on a criterion \cite{taylor,liu2017learning,luo2017thinet,elkerdawy2020filter,he2017channel,he2018soft}. Earlier filter-pruning methods \cite{luo2017thinet,Li2016PruningFF} require calculation of layer-wise sensitivity analysis to generate the model signature (i.e number of filters per layer) before pruning. Sensitivity analysis is computationally expensive, especially as models grow deeper. Recent methods \cite{taylor,slimming,wen2016learning} learn a global importance measure. Molchanov \etal \cite{taylor} propose a Taylor approximation on network's weights where the filter's gradients and norm are used to approximate its global importance score. Liu \etal \cite{slimming} and Wen \etal \cite{wen2016learning} introduce a sparsity loss in addition to the task loss as a regularization then prune filters whose criterion are less than a threshold. 

\textbf{Dynamic Pruning.} In contrast to one-model-fits-all deployment as in static pruning, dynamic pruning processes different routes per input sample. Similar to static pruning, methods can adopt different granularity to prune. Channel gating network (CGNet) \cite{hua2018channel} is a fine-grained method that skips zero locations in feature maps. A subset of input channels are processed by each kernel. The decision gating is learnt through regularization with a complex approximating non-differentiable function. They adopt group convolution and shuffle operation to balance filters frequency updates from different group of features. Closest to our work, we focus on dynamic filter pruning methods. In Runtime Neural Pruning (RNP) \cite{lin2017runtime}, a decision unit is modeled as a global recurrent layer, which generates discrete actions corresponding to four preset channel selection groups. The group selection is trained with reinforcement learning. Similarly, in BlockDrop \cite{wu2018blockdrop}, a policy network is trained to skip blocks in residual networks instead of only channels. D$^2$NN \cite{liu2018dynamic} defines a variant set of conditional branches in DNN, and uses Q-learning to train the branching policies. These methods train their policy functions by reinforcement learning which can be a non-trivial costly optimization task along with the CNN backbone. Feature Boosting and Suppression (FBS) \cite{FBS} method generates continuous channel saliency and uses a predefined pruning ratio per layer to obtain a discrete binary gating. LCS \cite{wang2020dynamic} propose to obtain a discrete action from N learned group of channels sampled from Gumbel distribution. They adopt annealing temperature to stabilize training and introduce diversity in the learned routes by gradient normalization trick. These existing regularization-based methods require additional careful tuning to stabilize training that is either caused by policy gradients or to enforce learning diverse routes as to not converge to a static pruning. Hence, we propose to formulate the channel selection as a self-supervised binary classification in which interpretable routes can be studied and simply trained with common SGD.

\section{Methodology}
In this section we first explain the mechanism for dynamic gating. Then, we discuss how we design the decision heads and the supervised loss

\subsection{Channel Gating}
Let $\boldsymbol{I^l}$, $\boldsymbol{W^l}$ be the input features, and weights of a convolution layer $l$, respectively, where $\boldsymbol{I^l} \in \mathbb{R}^{c_{l-1} \times w_l \times h_l}$, $\boldsymbol{W^l} \in \mathbb{R}^{c_l
\times c_{l-1} \times k_l \times k_l}$, and $c_l$ is the number of filters in layer $l$. A typical CNN block consists of a convolution operation ($*$), batch normalization ($\rm BN$), and an activation function ($f$) such as the commonly used ReLU. Without loss of generality, we ignore the bias term because of $\rm BN$ inclusion, thus, the output feature map $\boldsymbol{O^l}$ can be written as $\boldsymbol{O^l} = f({\rm BN}(\boldsymbol{I^l} * \boldsymbol{W^l}))$. We predict a binary mask $\boldsymbol{M^l} \in \mathbb{R}^{c_l}$ denoting the highly activated output feature maps $\boldsymbol{O^l}$ from the input activation map $\boldsymbol{I^l}$ by applying a decision head $f_p^l$ with learnable parameters $\boldsymbol{W_p^l}$. Masked output $\boldsymbol{I^{l+1}}$ is then represented as $\boldsymbol{I^{l+1}} = \boldsymbol{O^l} \odot Binarize(f_p^l(\boldsymbol{I^l};\boldsymbol{W_p}))$. $Binarize(.)$ function is $round(Sigmoid(.))$ to convert logits to a binary mask. The prediction of the highly activated output feature maps allows for processing filters $f$ where $M_f^l=1$ in the inference time and skipping the rest. Our decision head has $c_{l-1} \times c_{l}$ FLOPs cost per layer $l$ which is negligible.

\subsection{Self-Supervised Binary Gating}
Our proposed method as shown in Figure \ref{fig:dynamic} learns this dynamic routing in a self-supervised way by inserting a predictor head after each convolutional block to predict the highly $k$ activated filters of the next layer. The $k$ value is automatically calculated per input based on the mass of heatmap. 
\paragraph{Loss function} The ground truth binary mask of the highly activated features is attainable by sorting the norm of the features. The overall training objective is:

\begin{equation}
\begin{split}
\min\limits_{\{\boldsymbol{W},\boldsymbol{W_p}\}} L_{total} = & L_{ent} (f_{n}(\boldsymbol{x};\boldsymbol{W}), \boldsymbol{y}_k) \\
 & + L_{pred} (\{f_{p}^l(\boldsymbol{I}^l;\boldsymbol{W_p}^l), \boldsymbol{g}^l\}_L)
\end{split}
\end{equation}

where $f_{n}$ is the backbone of the baseline model, $L_{ent}$ is the cross-entropy task loss, $L_{pred}$ is the total predictor loss for all layers $l \in {1...L}$. In details, we define $L_{pred}$ as follows:
\begin{equation}
\begin{split}
& L_{pred} (\{\boldsymbol{P}^l, \boldsymbol{g}^l\}_L) = \\
& \sum_{l}^{L} \sum_{f}^{F_l} BCEWithLogits(\boldsymbol{P}^{l}_{f}, \boldsymbol{g}^{l}_{f}) 
\end{split}
\end{equation}

where $\boldsymbol{P}^l$ is the output of the decision head $f_{p}^l(\boldsymbol{I}^l;\boldsymbol{W_p}^l)$, $\boldsymbol{g^l}$ is the generated ground truth mask based on the top-$k$ highly activated output $\boldsymbol{O^l}$, $BCEWithLogits$ is a $Sigmoid$ followed by the binary cross entropy loss $BCE(p,g) = -{(g\log(p) + (1 - g)\log(1 - p))}$. 

\paragraph{Number of activated $\boldsymbol{k}$} We automatically calculate $k$ by keeping a constant percentage $r$ of the mass of heatmap. For each channel $i=1,...,c_l$, we keep the maximum response by applying a global maximum pooling (GMP) $(\text{GMP}{(\abs*{O^l_i})}$. For each input example, $k$ is the number of filters kept such that the cumulative mass of the sorted normalize activation reaches $r\%$. Ground truth generation algorithm is shown in Algorithm \ref{algo:gt}. We use the same $r$ for all layers, however, each sample will have different pruning ratio per layer based on its activation. As the target binary groundtruth is generated from the activations of the unmasked filters, FLOPs reduction can be loosely estimated prior to training to adjust $r$ accordingly. This advantage adds to the practicality of our method which does not rely on indirect hyperparameter tuning to reach a budget target in FLOPs reduction. It is also worth mentioning that $r=1$ is a special case that indicates the decision head will predict the completely deactivated features (e.g maximum response is zero). This enables maintaining accuracy of baseline with ideally trained decision head for highly sparse backbones.

\begin{algorithm}
\caption{Binary mask ground truth generation}
\label{algo:gt}
\begin{algorithmic}[1]
\Require{$I^{1} \dots I^{L}$, $r$} 
\Ensure{$g$ binary ground truth with 0 as to prune}
\State{$gt \gets ones(L, c_l)$}
\For{$l \gets 1$ to $L$}{
    \State {$acts$ $\gets$ {$\Call{GMP}{(\abs*{O^{l}})}$}}
    \State {$normalized$ $\gets$ {$acts/\sum{acts}$}}
    \State {$sorted, idx$ $\gets$ {$\Call{sort}{normalized, ``descend"}$}}
    \State {$cumulative$ $\gets$ {$\Call{cumsum}{sorted}$}}
    \State {$prune\_idx$ $\gets$ {$\Call{where}{cumulative > r}$}}
    \State {$gt[l][prune\_idx] \gets \mathbf{0}$ }
\EndFor
}
\end{algorithmic}
\end{algorithm}

\begin{table*}[t]
\centering
\begin{tabular}{clccc}
\hline
\multicolumn{1}{l}{} & Model & Dynamic? & Top-1 Acc. (\%) & FLOPs red. (\%) \\ \hline
\multirow{12}{*}{VGG16-BN} & Baseline & -- & 93.82 & --- \\
 & L1-norm \cite{li2016pruning} & N & 93.00 & 34 \\
 & ThiNet \cite{luo2017thinet} & N & 93.36 & 50 \\
 & CP \cite{he2017channel} & N & 93.18 & 50 \\
 & Taylor-50 \cite{taylor} & N & 92.00 & 51 \\
 & RNP \cite{lin2017runtime} & Y & 92.65 & 50 \\
 & FBS \cite{FBS} & Y & 93.03 & 50 \\
 & LCS \cite{wang2020dynamic}  & Y & 93.45 & 50 \\
 & \textbf{FTWT\textsubscript{J}} ($r=0.92$) & Y & 93.55 & \textbf{65} \\
  & \textbf{FTWT\textsubscript{D}} ($r=0.92$) & Y & \textbf{93.73} & 56 \\ \cline{2-5} 
  & Taylor-59 \cite{taylor} & N & 91.50 & 59 \\
 & \textbf{FTWT\textsubscript{D}} ($r=0.85$) & Y & \textbf{93.19} & 73 \\ 
  & \textbf{FTWT\textsubscript{J}} ($r=0.88$) & Y & 92.65 & \textbf{74} \\ \hline
\multicolumn{1}{l}{\multirow{7}{*}{ResNet56}} & Baseline & -- & 93.66 & -- \\
\multicolumn{1}{l}{} & Uniform from \cite{wang2020dynamic} & N & 74.39 & 50 \\
\multicolumn{1}{l}{} & ThiNet \cite{luo2017thinet} & N & 91.98 & 50 \\
\multicolumn{1}{l}{} & SFP \cite{he2018soft} & N & 92.56 & 48 \\
\multicolumn{1}{l}{} & LCS \cite{wang2020dynamic} & Y & 92.57 & 52 \\
\multicolumn{1}{l}{} & \textbf{FTWT\textsubscript{D}} ($r=0.80$) & Y & \textbf{92.63} & \textbf{66} \\ 
\multicolumn{1}{l}{} & \textbf{FTWT\textsubscript{J}} ($r=0.88$) & Y & 92.28 & 54 \\ \hline
\multicolumn{1}{l}{\multirow{5}{*}{MobileNetV1}} & Baseline & -- & 90.89 & -- \\
\multicolumn{1}{l}{} & MobileNet\_75 \cite{howard2017mobilenets} & N & 89.79 & 42 \\
\multicolumn{1}{l}{} & MobileNet\_50 \cite{howard2017mobilenets} & N & 87.58 & 73 \\
\multicolumn{1}{l}{} &  \textbf{FTWT\textsubscript{D}} ($r=1.0$) & Y & 91.06 & \textbf{78} \\ 
\multicolumn{1}{l}{} &  \textbf{FTWT\textsubscript{J}} ($r=1.0$) & Y & \textbf{91.21} & \textbf{78} \\ \hline
\end{tabular} 
\caption{Results on CIFAR-10. FLOPs red. indicates reduction in FLOPs in percentage. $r$ in our method states the hyperparameter ratio in Algorithm \ref{algo:gt}. \textbf{x} in FTWT\textsubscript{\textbf{x}} indicates joint (J) or decoupled (D) training.}
\label{tab:cifar10}
\end{table*}

\subsection{Prediction Head Design}
Prediction head design should be modeled in a simple way to reduce overhead over baseline network. In forward pass, we apply GMP that reduces feature map $\mathbf{I^l}$ per layer to $ E^l \in \mathbb{R}^{c^{l-1} \times 1 \times 1 }$. Next, we apply 1x1 convolution on the flattened embedding $E^l$ to produce the mask's logits. We experiment with two training modes: 1) decoupled, and 2) joint. In both modes, we train backbone weights and those of decision heads in parallel. The distinction is whether we do a fully differentiable training (joint) or stop gradients from heads to backpropagate to the backbone and vise versa (decoupled). In joint training, the decision head is fully differentiable except at the binarization part. Similar to previous works \cite{ye2020accelerating,elkerdawy2019lightweight,mallya2018piggyback}, we utilize straight-through estimator (STE) to bypass the non-differentiable function. An issue to consider is losses interference with multiple losses at different depth in the network as pointed out in \cite{huang2017multi}. Losses interference highlights that feature maps can be biased towards achieving high accuracy to local task more than the overall architecture. Unlike other methods that rely on careful training tuning to manage gradients from different losses, we train the heads along with backbone in parallel yet collaboratively as the masks are generated from the current status of the model. The groundtruth binary masks are explicitly adjusted by the updated backbone weights, thus, implicitly complying to the backbone learning speed. 

\section{Experiments and Analysis}
\label{sec:exp}

We evaluate our method on CIFAR \cite{cifar} and ImageNet \cite{deng2009imagenet} datasets on a variety of architectures such as VGG \cite{vgg}, ResNet \cite{resnet} and MobileNet \cite{howard2017mobilenets}. In all architectures, ground truth masks are generated after each conv-BN-ReLU block. For CIFAR baseline models, we train for 200 epochs using a batch-size of 128 with SGD optimizer. The initial learning rate (lr) $0.1$ is divided by 10 at epochs 80, 120, and 150. We use a momentum of 0.9 with weight decay of $5^{−4}$. For ImageNet, we use the pre-trained models in PyTorch \cite{pytorch} as baselines. Weights of decision heads are trained with $0.1$ as initial lr and same lr schedule as backbone. We use a 4 V100-GPU machine in our experiments. 

\begin{figure*}%
\centering
\begin{subfigure}[b]{0.95\columnwidth}
\includegraphics[width=\columnwidth]{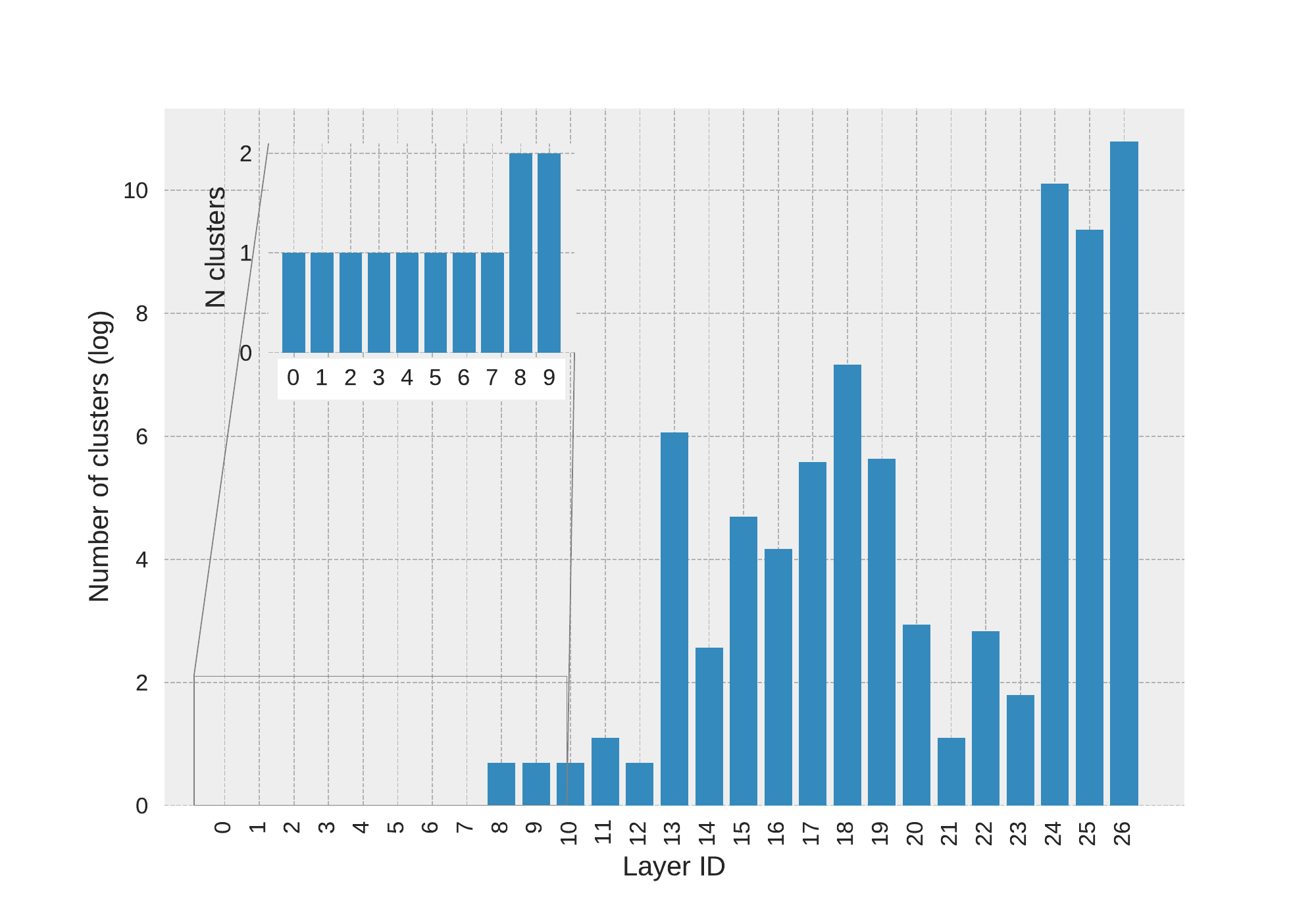}%
\caption{Number of unique group of filters (clusters) per layer.}%
\label{fig:mbnetcifara}%
\end{subfigure}%
\begin{subfigure}[b]{0.95\columnwidth}
\includegraphics[width=\columnwidth]{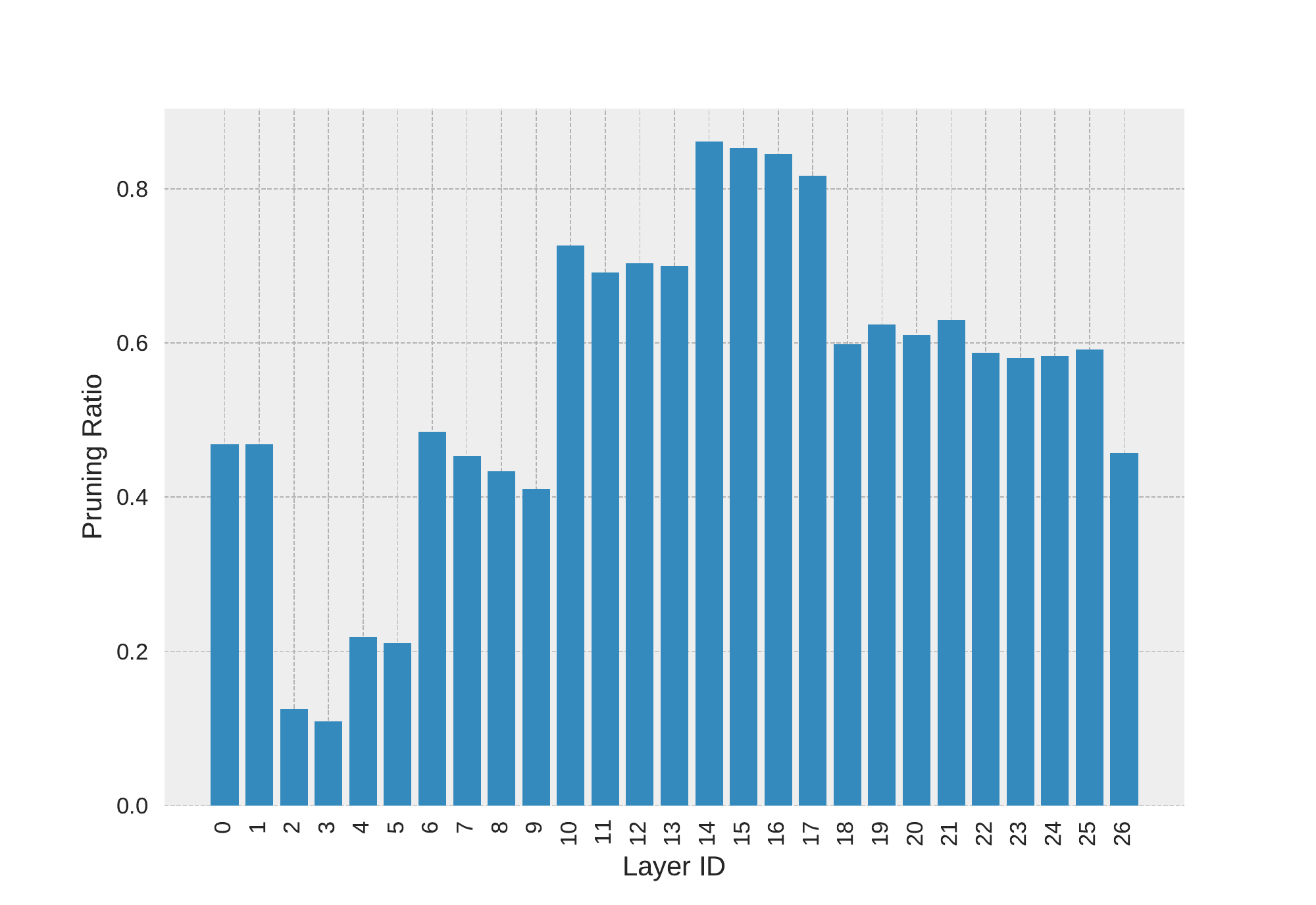}%
\caption{Pruning ratio per layer.}%
\label{fig:mbnetcifarb}%
\end{subfigure}\hfill%

\caption{MobileNetV1 CIFAR10 distributions}
\label{fig:mbnetcifar}
\end{figure*}

\begin{table*}[t]
\centering
\begin{tabular}{llccclc}
\hline
 & \multicolumn{1}{c}{Method} & Dynamic? & \multicolumn{3}{c}{Top-1 Acc. (\%)} & FLOPs red. (\%) \\
\multicolumn{3}{l}{} & \multicolumn{1}{l}{Baseline} & \multicolumn{1}{l}{Pruned} & Delta & \multicolumn{1}{l}{} \\ \hline
 
  \multirow{9}{*}{ResNet34} & Taylor \cite{taylor} & N & 73.31 & 72.83 & 0.48 & 22.25 \\ 
 & LCCL \cite{dong2017more} & Y & 73.42 & 72.99 & 0.43 & 24.80 \\
 & \textbf{FTWT ($r=0.97$)} & Y & 73.30 & 73.25 & \textbf{0.05} & 25.86 \\ 
  & \textbf{FTWT ($r=0.95$)} & Y & 73.30 & 72.79 & 0.51 & 37.77 \\ \cline{2-7} 
 & SFP \cite{he2018soft} & N & 73.92 & 71.83 & 2.09 & 41.10 \\
 & FPGM \cite{he2019filter} & N & 73.92 & 72.54 & 1.38 & 41.10 \\
 & \textbf{FTWT ($r=0.93$)} & Y & 73.30 & 72.17 & \textbf{1.13} & 47.42 \\
 \cline{2-7}
 & ResNet18 \cite{resnet}& N & 73.30 & 69.76 & 3.54 & 50.04 \\
 & \textbf{FTWT ($r=0.92$)} & Y & 73.30 & 71.71 & \textbf{1.59
} & 52.24 \\\hline
 
\multirow{5}{*}{ResNet18} 
 & PFP-B \cite{LiebenweinBLFR20} & N & 69.74 & 65.65 & 4.09 & 43.12 \\
 & SFP \cite{he2018soft} & N & 70.28 & 67.10 & 3.18 & 41.80 \\
 & LCCL \cite{dong2017more} & Y & 69.98 & 66.33 & 3.65 & 34.60 \\
 & FBS \cite{FBS} & Y & 70.70 & 68.20 & 2.50 & 49.49 \\
 & \textbf{FTWT ($r=0.91$)} & Y & 69.76 & 67.49 & \textbf{2.27} & 51.56 \\ \hline

 \multirow{2}{*}{MobileNetV1} & MobileNetV1-75 \cite{howard2017mobilenets} & N & 69.76 & 67.00 & 2.76 & 42.85 \\
 & \textbf{FTWT ($r=1$)} & Y & 69.57 & 69.66 & \textbf{-0.09} & 41.07 \\ \hline
\end{tabular}
\caption{Results on ImageNet. Baseline accuracy for each method is reported along with the pruned model's accuracy and accuracy change from baseline. FLOPs red. represents reduction in FLOPs in percentage. Negative delta indicates increase in accuracy from the baseline. $r$ in our method states the hyperparameter ratio in Algorithm \ref{algo:gt}}
\label{tab:imagenet}
\end{table*}
\subsection{Experiments on CIFAR}
\label{sec:cifarsec}
We follow similar training settings used in baseline for dynamic training, but we train all models with initial learning rate of $1e^{-2}$. We report the average accuracy over three repeated experiments and FLOPs reduction on CIFAR-10 on multiple architectures in Table \ref{tab:cifar10}. Our method (FTWT) achieves higher FLOPs reduction on similar top1-accuracy than static and dynamic pruning methods. We achieve up to 66\% FLOPs reduction on VGG-16 and ResNet-56, that is higher than dynamic filter pruning methods RNP \cite{lin2017runtime}, FBS\cite{FBS}, LCS \cite{wang2020dynamic} by up to 15\%. Joint training performs equally well as decoupled training on high $r$ thresholds. However, the accuracy drops in comparison to decoupled training on lower thresholds. That is due to conflict increase between losses as can be seen on $\approx$ 73\% FLOPs reduction on VGG. We further achieve 73\% FLOPs reduction on VGG with only 0.63\% accuracy drop. Moreover, FTWT outperforms smaller variants of MobileNet in accuracy by 3.42\% with higher FLOPs reduction. 

We visualize the number of unique combination of filters (clusters) that are activated over the whole dataset $D$ and the pruning ratio per layer in Figure \ref{fig:mbnetcifar}. Meaning that, each sample $i$ that produces a binary mask $m_{i,j}$ per layer $j$, the unique clusters per layer is $set({m_{0,j},...,m_{i,j},...m_{|D|,j}})$. In LCS and RNP, a fixed number of clusters is preset as a hyperparameter for all layers, we show in Figure \ref{fig:mbnetcifara} that layers differ in the number of diverse clusters. Our method adjusts different number of clusters per layer automatically due to self-supervised mask generation mechanism. For easier visualization, y-axis is shown in log scale. Early layers have small diversity in the group of filters activated, thus, act similar to static pruning. This is sensible as early layers detect low-level features and have less dependency on the input. On the other hand, the number of clusters increases as we go deeper in the network. It is worth mentioning that these different clusters are fine-grained, which means clusters can differ in one filter only. We also calculated the percentage of core filters that are shared among all clusters per layer. We found that the range of the percentage of core filters with respect to total filters vary from 0.4 to 1.0. This gives insight on why static pruning methods result in large drop in accuracy with large pruning. As the attainable pruning ratio is limited by the number of core filters and further pruning will limit the model's capacity. An interesting future research question would be if we can determine the compressibility of a model based on the core filters ratio notion. Finally, Figure \ref{fig:mbnetcifarb} shows the pruning ratio per layer, as to be expected, the later layers are more heavily pruned than early layers as layers get wider and more compressible. We notice heavy pruning reaching 85\% in the middle layers with a sequence of layers with 512 filters.

\subsection{Experiments on ImageNet}
For ImageNet, we train for 90 epochs with initial learning rate of $10^{-2}$ that decays each 30 epochs by 0.1. Experiments on ImageNet is done with the decoupled training mode. Table \ref{tab:imagenet} shows drop in accuracy from baseline for each method to account for training differences due to augmentations. Results show that our method achieve smaller drop in accuracy with higher FLOPs reduction in comparison to other SOTA methods. We achieve similar accuracy reduction as LCCL with 13\% higher FLOPs reduction on ResNet34. On the other hand, on similar FLOPs reduction ($\approx 25$), we have minimal drop in accuracy ($\approx 0.05\%$). Althought we achieve similar accuracy on ResNet18 with FBS, the latter requires a predefined number of filters to keep per layer. On the other hand, our method dynamically assigns pruning ratio per layer which shows the effectiveness of our heatmap mass as a criteria. We also compare with architecture's smaller variants such as ResNet18 and MobileNet-75. We outperform ResNet18 and MobileNet-75 by $\approx 2\%$ in accuracy on similar computation budget. 

\subsection{Ablation Study}

\begin{table*}[!htbp]
\centering
\begin{subtable}[]{\columnwidth}
\centering
\begin{tabular}{|c|c||c|c|c|}
\hline
$\sigma$ & \begin{tabular}[c]{@{}c@{}}Dense\\ model\end{tabular} & \begin{tabular}[c]{@{}c@{}}FTWT\\ (ours)\end{tabular} & \begin{tabular}[c]{@{}c@{}}Taylor\\ Pruning\end{tabular} & \begin{tabular}[c]{@{}c@{}}Uniform\\ Pruning\end{tabular} \\ \hline
0.5 & 0.11 & \textbf{0.12} & 0.20 & \textbf{0.12 }\\ \hline
0.7 & 0.16 & \textbf{0.18} & 0.39 & 0.19 \\ \hline
0.9 & 0.38 & \textbf{0.39} & 0.57 & 0.42 \\ \hline
1.09 & 0.69 & \textbf{0.58} & 0.66 & 0.61 \\ \hline
1.27 & 0.74 & \textbf{0.68} & 0.69 & 0.71 \\ \hline
1.45 & 0.76 & \textbf{0.73} & 0.74 & 0.75 \\ \hline
\end{tabular}
\caption{Gaussian blur noise.}
\label{tab:blurgaussian}
\end{subtable}
\begin{subtable}[]{\columnwidth}
\centering
\begin{tabular}{|c|c||c|c|c|}
\hline
$\sigma$ & \begin{tabular}[c]{@{}c@{}}Dense\\ model\end{tabular} & \begin{tabular}[c]{@{}c@{}}FTWT\\ (ours)\end{tabular} & \begin{tabular}[c]{@{}c@{}}Taylor\\ Pruning\end{tabular} & \begin{tabular}[c]{@{}c@{}}Uniform\\ Pruning\end{tabular} \\ \hline
0.00 & 0.11 & \textbf{0.12} & 0.16 & \textbf{0.12} \\ \hline
0.02 & 0.11 & \textbf{0.12} & 0.16 & \textbf{0.12} \\ \hline
0.05 & 0.11 & \textbf{0.12} & 0.17 & 0.13 \\ \hline
0.11 & 0.13 & \textbf{0.14} & 0.20 & 0.15 \\ \hline
0.14 & 0.19 & \textbf{0.21} & 0.30 & 0.22 \\ \hline
0.20 & 0.39 & 0.43 & 0.51 & \textbf{0.42} \\ \hline
\end{tabular}
\caption{Additive Gaussian noise.}
\label{tab:addgaussian}
\end{subtable}

\begin{subtable}[]{2\columnwidth}
\centering
\begin{tabular}{|c|c|c|c|}
\hline
\multicolumn{2}{|c|}{Gaussian Blur} & \multicolumn{2}{c|}{Additive Noise} \\ \hline
\begin{tabular}[c]{@{}c@{}}FTWT\\  with normalization\end{tabular} & \begin{tabular}[c]{@{}c@{}}FTWT \\ without normalization\end{tabular} & \begin{tabular}[c]{@{}c@{}}FTWT\\  with normalization\end{tabular} & \begin{tabular}[c]{@{}c@{}}FTWT\\  without normalization\end{tabular} \\ \hline
\textbf{0.12} & 0.19 & \textbf{0.12} & 0.12 \\ \hline
\textbf{0.18} & 0.48 & \textbf{0.12} & 0.13 \\ \hline
\textbf{0.39} & 0.73 & \textbf{0.14} & 0.15 \\ \hline
\textbf{0.58} & 0.85 & \textbf{0.21} & 0.24 \\ \hline
\textbf{0.68} & 0.90 & \textbf{0.43} & 0.47 \\ \hline
\end{tabular}
\caption{Our method with and without softmax normalization in decision heads.}
\label{tab:softmax}
\end{subtable}

\caption{Dataset shift experiments: Numbers represent Brier score on CIFAR-10 VGG16}
\end{table*}

\subsubsection{Uncertainty under Dataset Shift}
In this section, we measure the sensitivity of our routing to dataset shift. Metrics under dataset shift are rarely inspected in model pruning literature. We believe in its importance as inference complexity increases and thus would like to initiate reporting such comparisons. We conduct experiments for VGG16-bn CIFAR-10 with a high pruning ratio 73\% for all pruned models. Inspired by \cite{ovadia2019can}, we report Brier score \cite{brier1950verification} under different type of noise such as Gaussian blur Table \ref{tab:blurgaussian} and additive noise Table \ref{tab:addgaussian} with baseline dense model as a reference. As can be seen, our method is more resilient than static Taylor pruning with a lower brier scores. We also compare with static uniform pruning, we achieve a similar (sometimes slightly lower) Brier score. This shows the resilience of our model to data shift even when compared with static pruning decision that is not data-dependent. Finally, as to be expected, the dense model is the most resilient to noise. However, our method still shows a fair quality matching overall. We attribute this distribution stability to the softmax in the head. The softmax acts as a normalizer which reduces sensitivity to distribution shift. We compare our method with and without softmax normalization in the decision head to verify this hypothesis. Table \ref{tab:softmax} shows Brier scores with additive and blurring noise for this comparison. As can be seen, indeed, the normalizer stabilizes the decision masks output especially in the case of blurring. 


\subsubsection{Dynamic Signature and Dynamic Routing}
\label{sec:predef}
We investigate decoupling the effect from the dynamic signature (i.e. pruning ratio per layer) per sample from the dynamic routing (i.e group of filters to be activated). We explore the effectiveness of dynamic routing with a pre-defined signature for all inputs. In these experiments, signature is pre-defined using Taylor criteria proposed in \cite{taylor} as a case study. As in previous setup, we select the highly activated k features where k is defined by the signature while samples differ in which k filters are selected. Table \ref{tab:defined} shows results of dynamic routing under different pruning ratios. As can be seen, dynamic routing performs better than static inference especially on high pruning ratio by up to 4\%. Training setup is the same for CIFAR as explained in paper, however, we train ImageNet models for 30 epochs using finetune setup instead of training setup with 90 epochs as differentiated in \cite{liu2018rethinking} to accelerate the experiment.

\subsubsection{Hyperparameter $r$ selection}
The hyperparameter $r$ (i.e mass ratio) is selected based on a simple evaluation before training. W estimate the FLOPs before training by applying the groundtruth masks on the pretrained frozen dense model over the training set (one-shot pass). Subsequently, we get an estimate before training is initiated of the expected FLOPs reduction under different r values. This simplifies hyperparameter selection to achieve a target FLOPs reduction. On the other hand, sparse regularization hyperparameter is usually fine-tuned with a cross-validation process and requires a trial and error of multiple full training to achieve a target budget. There is no direct relation between the regularization weight and the final achieved FLOPs reduction knowingly before training. Our method simplifies the selection and makes it a more practical option when a target budget is given as a prior. Table \ref{tab:flops} shows the estimated FLOPs before training using different thresholds, $r$, and the actual reached FLOPs reduction after training. The difference in reduction is due to the inaccuracy of the decision heads. Nonetheless, the estimated FLOPs gives a good approximation to the final reached FLOPs and thus reduce the hyperparameter search. 

\begin{table}[t]
\centering
\begin{tabular}{l|c|c}
\hline
Model & \begin{tabular}[c]{@{}c@{}}Est. \\ FLOPs (\%)\end{tabular} & \begin{tabular}[c]{@{}c@{}}Final\\ FLOPs (\%)\end{tabular} \\ \hline
$\text{MobileNet}_{(1.0)}$ & 42.3  & 41.07 \\ \hline
$\text{Resnet34}_{(0.97)}$ & 23.32 & 25.86 \\
$\text{Resnet34}_{(0.95)}$ & 31.77 & 37.77 \\ \hline
\end{tabular}

\caption{Estimated FLOPs \textit{before} training under different thresholds (indicated in parentheses) vs achieved FLOPs after training.}
\label{tab:flops}
\end{table}

\begin{table}[t]
\centering
\begin{tabular}{l|c|c}
\hline
Model & \begin{tabular}[c]{@{}c@{}}FLOPs\\  reduction (\%)\end{tabular} & \begin{tabular}[c]{@{}c@{}}Latency \\ reduction (\%)\end{tabular} \\ \hline
\multirow{3}{*}{ResNet34} & 52.18 & 27.17 \\
                          & 37.77 & 19.78 \\
                          & 25.86 & 11.08  \\ \hline
\end{tabular}
\caption{Realistic vs theoretical speedup on ImageNet on AMD Ryzen Threadripper 2970WX CPU with batch size of 1.}
\label{tab:latency}
\end{table}

\subsection{Theoretical vs Practical Speedup}
\label{sec:limit}
For all compression methods, including static and dynamic pruning, there is often a wide gap between FLOPs reduction and realistic speedup due to other factors such as I/O delays and BLAS libraries. Speedup is hardware and backend dependent as shown in prior works \cite{elkerdawy2020filter,yang2017designing,bianco2018benchmark}. We test the realistic speed on PyTorch \cite{pytorch} using MKL backend on AMD CPU using a single thread as shown on Table \ref{tab:latency}.

\textbf{Limitations.} Our realistic speedup is less than FLOPs reduction and that is attributed to two factors: 1) Data transfer overhead from slicing the dense weight matrix based on the mask prediction, which can be mitigated by backends with efficient in-place sparse inference. 2) Speed up is dependent on the model's signature and hardware's specs. Pruning from later layers that process smaller input resolution might not achieve as much speedup as pruning from early layers. Constraint aware optimization using Alternating Direction Method of Multipliers (ADMM) \cite{admm} such as proposed in \cite{yang2019ecc} can be further integrated with our method to optimize over latency instead of FLOPs.

\begin{table}[!htbp]
\centering
\begin{tabular}{l|l|ccc}
\hline
\multirow{2}{*}{Dataset} & \multirow{2}{*}{Model} & \multirow{2}{*}{\begin{tabular}[c]{@{}c@{}}FLOPs \\ (\%)\end{tabular}} & \multicolumn{2}{c}{Top-1 acc. (\%)} \\
 &  &  & Static & Dynamic \\ \hline
\multirow{3}{*}{CIFAR-10} & \multirow{2}{*}{VGG16-BN} & 50 & 92.00 & \textbf{93.80} \\
 &  & 85 & 91.12 & \textbf{92.75} \\ \cline{2-5} 
 & ResNet56 & 70 & 91.61 & \textbf{92.09} \\ \hline
\multirow{3}{*}{CIFAR-100} & \multirow{3}{*}{VGG16-BN} & 30 & 72.65 & \textbf{73.67} \\
 &  & 65 & 68.17 & \textbf{72.18} \\
 &  & 93 & 58.74 & \textbf{60.05} \\ \hline
ImageNet & ResNet18 & 45 & 64.89 & \textbf{65.11} \\ \hline
\end{tabular}
\caption{Accuracy comparison of dynamic routing with a pre-defined signature and its counterpart with static inference.}
\label{tab:defined}
\end{table}

\section{Conclusion}
In this paper, we propose a novel formulation for dynamic model pruning. Similar to other dynamic pruning methods, we equip a cheap decision head to the original convolutional layer. However, we propose to train the decision heads in a self-supervised paradigm. This head predicts the most likely to be highly activated filters given the layer's input activation. The masks are trained using a binary cross entropy loss decoupled from the task loss to remove losses interference. We generate the mask ground truth based on a novel criteria using the heatmap mass per input sample. In our experiments, we showed results on various architectures on CIFAR and ImageNet datasets, and our approach outperforms other dynamic and static pruning methods under similar FLOPs reduction.
\section*{Acknowledgment}
We thank the reviewers for their valuable feedback. We also would like to thank Compute Canada for their supercomputers to conduct our experiments.

{\small
\bibliographystyle{ieee_fullname}
\bibliography{egbib}

\begin{thebibliography}{10}\itemsep=-1pt

\bibitem{bianco2018benchmark}
Simone Bianco, Remi Cadene, Luigi Celona, and Paolo Napoletano.
\newblock Benchmark analysis of representative deep neural network
  architectures.
\newblock {\em IEEE Access}, 6:64270--64277, 2018.

\bibitem{admm}
Stephen Boyd, Neal Parikh, and Eric Chu.
\newblock {\em Distributed optimization and statistical learning via the
  alternating direction method of multipliers}.
\newblock Now Publishers Inc, 2011.

\bibitem{brier1950verification}
Glenn~W Brier et~al.
\newblock Verification of forecasts expressed in terms of probability.
\newblock {\em Monthly weather review}, 78(1):1--3, 1950.

\bibitem{cai2019once}
Han Cai, Chuang Gan, Tianzhe Wang, Zhekai Zhang, and Song Han.
\newblock Once-for-all: Train one network and specialize it for efficient
  deployment.
\newblock {\em arXiv preprint arXiv:1908.09791}, 2019.

\bibitem{deng2009imagenet}
Jia Deng, Wei Dong, Richard Socher, Li-Jia Li, Kai Li, and Li Fei-Fei.
\newblock Imagenet: A large-scale hierarchical image database.
\newblock In {\em 2009 IEEE conference on computer vision and pattern
  recognition}, pages 248--255. Ieee, 2009.

\bibitem{dong2017more}
Xuanyi Dong, Junshi Huang, Yi Yang, and Shuicheng Yan.
\newblock More is less: A more complicated network with less inference
  complexity.
\newblock In {\em Proceedings of the IEEE Conference on Computer Vision and
  Pattern Recognition}, pages 5840--5848, 2017.

\bibitem{elkerdawy2020filter}
Sara Elkerdawy, Mostafa Elhoushi, Abhineet Singh, Hong Zhang, and Nilanjan Ray.
\newblock To filter prune, or to layer prune, that is the question.
\newblock In {\em Proceedings of the Asian Conference on Computer Vision},
  2020.

\bibitem{elkerdawy2019lightweight}
Sara Elkerdawy, Hong Zhang, and Nilanjan Ray.
\newblock Lightweight monocular depth estimation model by joint end-to-end
  filter pruning.
\newblock In {\em 2019 IEEE International Conference on Image Processing
  (ICIP)}, pages 4290--4294. IEEE, 2019.

\bibitem{frankle2018lottery}
Jonathan Frankle and Michael Carbin.
\newblock The lottery ticket hypothesis: Finding sparse, trainable neural
  networks.
\newblock {\em arXiv preprint arXiv:1803.03635}, 2018.

\bibitem{FBS}
Xitong Gao, Yiren Zhao, {\L}ukasz Dudziak, Robert Mullins, and Cheng-zhong Xu.
\newblock Dynamic channel pruning: Feature boosting and suppression.
\newblock {\em arXiv preprint arXiv:1810.05331}, 2018.

\bibitem{han2015learning}
Song Han, Jeff Pool, John Tran, and William Dally.
\newblock Learning both weights and connections for efficient neural network.
\newblock In {\em Advances in Neural Information Processing Systems (NIPS)},
  pages 1135--1143, 2015.

\bibitem{resnet}
Kaiming He, Xiangyu Zhang, Shaoqing Ren, and Jian Sun.
\newblock Deep residual learning for image recognition.
\newblock In {\em Proceedings of the IEEE conference on computer vision and
  pattern recognition}, pages 770--778, 2016.

\bibitem{he2018soft}
Yang He, Guoliang Kang, Xuanyi Dong, Yanwei Fu, and Yi Yang.
\newblock Soft filter pruning for accelerating deep convolutional neural
  networks.
\newblock {\em arXiv preprint arXiv:1808.06866}, 2018.

\bibitem{he2019filter}
Yang He, Ping Liu, Ziwei Wang, Zhilan Hu, and Yi Yang.
\newblock Filter pruning via geometric median for deep convolutional neural
  networks acceleration.
\newblock In {\em Proceedings of the IEEE/CVF Conference on Computer Vision and
  Pattern Recognition}, pages 4340--4349, 2019.

\bibitem{he2017channel}
Yihui He, Xiangyu Zhang, and Jian Sun.
\newblock Channel pruning for accelerating very deep neural networks.
\newblock In {\em Proceedings of the IEEE International Conference on Computer
  Vision}, pages 1389--1397, 2017.

\bibitem{howard2017mobilenets}
Andrew~G Howard, Menglong Zhu, Bo Chen, Dmitry Kalenichenko, Weijun Wang,
  Tobias Weyand, Marco Andreetto, and Hartwig Adam.
\newblock Mobilenets: Efficient convolutional neural networks for mobile vision
  applications.
\newblock {\em arXiv preprint arXiv:1704.04861}, 2017.

\bibitem{hu2018squeeze}
Jie Hu, Li Shen, and Gang Sun.
\newblock Squeeze-and-excitation networks.
\newblock In {\em Proceedings of the IEEE conference on computer vision and
  pattern recognition}, pages 7132--7141, 2018.

\bibitem{hua2018channel}
Weizhe Hua, Yuan Zhou, Christopher De~Sa, Zhiru Zhang, and G~Edward Suh.
\newblock Channel gating neural networks.
\newblock {\em arXiv preprint arXiv:1805.12549}, 2018.

\bibitem{huang2017multi}
Gao Huang, Danlu Chen, Tianhong Li, Felix Wu, Laurens van~der Maaten, and
  Kilian~Q Weinberger.
\newblock Multi-scale dense networks for resource efficient image
  classification.
\newblock {\em arXiv preprint arXiv:1703.09844}, 2017.

\bibitem{cifar}
Alex Krizhevsky, Geoffrey Hinton, et~al.
\newblock Learning multiple layers of features from tiny images.
\newblock 2009.

\bibitem{li2020eagleeye}
Bailin Li, Bowen Wu, Jiang Su, and Guangrun Wang.
\newblock Eagleeye: Fast sub-net evaluation for efficient neural network
  pruning.
\newblock In {\em European conference on computer vision}, pages 639--654.
  Springer, 2020.

\bibitem{li2016pruning}
Hao Li, Asim Kadav, Igor Durdanovic, Hanan Samet, and Hans~Peter Graf.
\newblock Pruning filters for efficient convnets.
\newblock {\em arXiv preprint arXiv:1608.08710}, 2016.

\bibitem{Li2016PruningFF}
Hao Li, Asim Kadav, Igor Durdanovic, Hanan Samet, and Hans~Peter Graf.
\newblock Pruning filters for efficient convnets.
\newblock {\em ICLR}, 2017.

\bibitem{li2019additive}
Yuhang Li, Xin Dong, and Wei Wang.
\newblock Additive powers-of-two quantization: An efficient non-uniform
  discretization for neural networks.
\newblock {\em arXiv preprint arXiv:1909.13144}, 2019.

\bibitem{liebenwein2021lost}
Lucas Liebenwein, Cenk Baykal, Brandon Carter, David Gifford, and Daniela Rus.
\newblock Lost in pruning: The effects of pruning neural networks beyond test
  accuracy.
\newblock {\em arXiv preprint arXiv:2103.03014}, 2021.

\bibitem{LiebenweinBLFR20}
Lucas Liebenwein, Cenk Baykal, Harry Lang, Dan Feldman, and Daniela Rus.
\newblock Provable filter pruning for efficient neural networks.
\newblock In {\em 8th International Conference on Learning Representations,
  {ICLR} 2020, Addis Ababa, Ethiopia, April 26-30, 2020}. OpenReview.net, 2020.

\bibitem{lin2017runtime}
Ji Lin, Yongming Rao, Jiwen Lu, and Jie Zhou.
\newblock Runtime neural pruning.
\newblock In {\em Proceedings of the 31st International Conference on Neural
  Information Processing Systems}, pages 2178--2188, 2017.

\bibitem{liu2018dynamic}
Lanlan Liu and Jia Deng.
\newblock Dynamic deep neural networks: Optimizing accuracy-efficiency
  trade-offs by selective execution.
\newblock In {\em Proceedings of the AAAI Conference on Artificial
  Intelligence}, volume~32, 2018.

\bibitem{liu2017learning}
Zhuang Liu, Jianguo Li, Zhiqiang Shen, Gao Huang, Shoumeng Yan, and Changshui
  Zhang.
\newblock Learning efficient convolutional networks through network slimming.
\newblock In {\em Proceedings of the IEEE International Conference on Computer
  Vision}, pages 2736--2744, 2017.

\bibitem{slimming}
Zhuang Liu, Jianguo Li, Zhiqiang Shen, Gao Huang, Shoumeng Yan, and Changshui
  Zhang.
\newblock Learning efficient convolutional networks through network slimming.
\newblock In {\em Proceedings of the IEEE ICCV}, pages 2736--2744, 2017.

\bibitem{liu2018rethinking}
Zhuang Liu, Mingjie Sun, Tinghui Zhou, Gao Huang, and Trevor Darrell.
\newblock Rethinking the value of network pruning.
\newblock {\em arXiv preprint arXiv:1810.05270}, 2018.

\bibitem{louizos2017learning}
Christos Louizos, Max Welling, and Diederik~P Kingma.
\newblock Learning sparse neural networks through $ l\_0 $ regularization.
\newblock {\em arXiv preprint arXiv:1712.01312}, 2017.

\bibitem{lowel1992selection}
Siegrid Lowel and Wolf Singer.
\newblock Selection of intrinsic horizontal connections in the visual cortex by
  correlated neuronal activity.
\newblock {\em Science}, 255(5041):209--212, 1992.

\bibitem{luo2017thinet}
Jian-Hao Luo, Jianxin Wu, and Weiyao Lin.
\newblock Thinet: A filter level pruning method for deep neural network
  compression.
\newblock In {\em Proceedings of the IEEE international conference on computer
  vision}, pages 5058--5066, 2017.

\bibitem{mallya2018piggyback}
Arun Mallya, Dillon Davis, and Svetlana Lazebnik.
\newblock Piggyback: Adapting a single network to multiple tasks by learning to
  mask weights.
\newblock In {\em Proceedings of the European Conference on Computer Vision
  (ECCV)}, pages 67--82, 2018.

\bibitem{taylor}
Pavlo Molchanov, Arun Mallya, Stephen Tyree, Iuri Frosio, and Jan Kautz.
\newblock Importance estimation for neural network pruning.
\newblock In {\em Proceedings of the IEEE/CVF Conference on Computer Vision and
  Pattern Recognition}, pages 11264--11272, 2019.

\bibitem{ovadia2019can}
Yaniv Ovadia, Emily Fertig, Jie Ren, Zachary Nado, David Sculley, Sebastian
  Nowozin, Joshua~V Dillon, Balaji Lakshminarayanan, and Jasper Snoek.
\newblock Can you trust your model's uncertainty? evaluating predictive
  uncertainty under dataset shift.
\newblock {\em arXiv preprint arXiv:1906.02530}, 2019.

\bibitem{paganini2020prune}
Michela Paganini.
\newblock Prune responsibly.
\newblock {\em arXiv preprint arXiv:2009.09936}, 2020.

\bibitem{pytorch}
Adam Paszke, Sam Gross, Francisco Massa, Adam Lerer, James Bradbury, Gregory
  Chanan, Trevor Killeen, Zeming Lin, Natalia Gimelshein, Luca Antiga, et~al.
\newblock Pytorch: An imperative style, high-performance deep learning library.
\newblock {\em arXiv preprint arXiv:1912.01703}, 2019.

\bibitem{sharify2019laconic}
Sayeh Sharify, Alberto~Delmas Lascorz, Mostafa Mahmoud, Milos Nikolic, Kevin
  Siu, Dylan~Malone Stuart, Zissis Poulos, and Andreas Moshovos.
\newblock Laconic deep learning inference acceleration.
\newblock In {\em 2019 ACM/IEEE 46th Annual International Symposium on Computer
  Architecture (ISCA)}, pages 304--317. IEEE, 2019.

\bibitem{vgg}
Karen Simonyan and Andrew Zisserman.
\newblock Very deep convolutional networks for large-scale image recognition.
\newblock In Yoshua Bengio and Yann LeCun, editors, {\em 3rd International
  Conference on Learning Representations, {ICLR} 2015, San Diego, CA, USA, May
  7-9, 2015, Conference Track Proceedings}, 2015.

\bibitem{tan2019mnasnet}
Mingxing Tan, Bo Chen, Ruoming Pang, Vijay Vasudevan, Mark Sandler, Andrew
  Howard, and Quoc~V Le.
\newblock Mnasnet: Platform-aware neural architecture search for mobile.
\newblock In {\em Proceedings of the IEEE/CVF Conference on Computer Vision and
  Pattern Recognition}, pages 2820--2828, 2019.

\bibitem{tang2020scop}
Yehui Tang, Yunhe Wang, Yixing Xu, Dacheng Tao, Chunjing Xu, Chao Xu, and Chang
  Xu.
\newblock Scop: Scientific control for reliable neural network pruning.
\newblock {\em Advances in Neural Information Processing Systems},
  33:10936--10947, 2020.

\bibitem{wang2020dynamic}
Yulong Wang, Xiaolu Zhang, Xiaolin Hu, Bo Zhang, and Hang Su.
\newblock Dynamic network pruning with interpretable layerwise channel
  selection.
\newblock In {\em Proceedings of the AAAI Conference on Artificial
  Intelligence}, volume~34, pages 6299--6306, 2020.

\bibitem{wen2016learning}
Wei Wen, Chunpeng Wu, Yandan Wang, Yiran Chen, and Hai Li.
\newblock Learning structured sparsity in deep neural networks.
\newblock In {\em Advances in neural information processing systems}, pages
  2074--2082, 2016.

\bibitem{williams1992simple}
Ronald~J Williams.
\newblock Simple statistical gradient-following algorithms for connectionist
  reinforcement learning.
\newblock {\em Machine learning}, 8(3-4):229--256, 1992.

\bibitem{wu2018blockdrop}
Zuxuan Wu, Tushar Nagarajan, Abhishek Kumar, Steven Rennie, Larry~S Davis,
  Kristen Grauman, and Rogerio Feris.
\newblock Blockdrop: Dynamic inference paths in residual networks.
\newblock In {\em Proceedings of the IEEE Conference on Computer Vision and
  Pattern Recognition}, pages 8817--8826, 2018.

\bibitem{yang2019ecc}
Haichuan Yang, Yuhao Zhu, and Ji Liu.
\newblock Ecc: Platform-independent energy-constrained deep neural network
  compression via a bilinear regression model.
\newblock In {\em Proceedings of the IEEE/CVF Conference on Computer Vision and
  Pattern Recognition}, pages 11206--11215, 2019.

\bibitem{yang2017designing}
Tien-Ju Yang, Yu-Hsin Chen, and Vivienne Sze.
\newblock Designing energy-efficient convolutional neural networks using
  energy-aware pruning.
\newblock In {\em Proceedings of the IEEE Conference on Computer Vision and
  Pattern Recognition}, pages 5687--5695, 2017.

\bibitem{ye2020accelerating}
Xucheng Ye, Pengcheng Dai, Junyu Luo, Xin Guo, Yingjie Qi, Jianlei Yang, and
  Yiran Chen.
\newblock Accelerating cnn training by pruning activation gradients.
\newblock In {\em European Conference on Computer Vision}, pages 322--338.
  Springer, 2020.

\bibitem{zhang2018lq}
Dongqing Zhang, Jiaolong Yang, Dongqiangzi Ye, and Gang Hua.
\newblock Lq-nets: Learned quantization for highly accurate and compact deep
  neural networks.
\newblock In {\em Proceedings of the European conference on computer vision
  (ECCV)}, pages 365--382, 2018.

\bibitem{zhuang2018discrimination}
Zhuangwei Zhuang, Mingkui Tan, Bohan Zhuang, Jing Liu, Yong Guo, Qingyao Wu,
  Junzhou Huang, and Jinhui Zhu.
\newblock Discrimination-aware channel pruning for deep neural networks.
\newblock {\em Advances in neural information processing systems}, 31, 2018.

\end{thebibliography}
}

\beginsupplement

\section*{Appendix}

\section{Compute Information}
Table \ref{tab:compute} shows the compute hours for the training of the reported models in our paper using a 4 V100 GPU machine.

\begin{table}[!htbp]
\centering
\begin{tabular}{l|l|c}
\hline
Dataset & Model & \multicolumn{1}{l}{Compute hours} \\ \hline
\multicolumn{1}{c|}{
\multirow{3}{*}{CIFAR-10}} & VGG & 2.0 \\
 & ResNet56 & 8.0 \\
 & MobileNet & 3.5 \\ \hline
\multirow{3}{*}{ImageNet} & ResNet18 &  40.0 \\
 & ResNet34 & 55.5 \\ 
 & MobileNetv1 & 41.0 \\ \hline
\end{tabular}
\caption{Compute hours on a 4 V100 GPU machine}
\label{tab:compute}
\end{table}

\section{CIFAR}
\subsection{MobileNet}
We compare our method on MobileNetv1/v2 under different pruning ratio with other pruning methods such as EagleEye \cite{li2020eagleeye}, SCOP \cite{tang2020scop} and DCP \cite{zhuang2018discrimination}. Note that EagleEye reports the best out of two candidate models different in signature, thus double the training time. We outperform SOTA by 37\% higher FLOPs reduction on similar accuracy as shown in \ref{fig:resnet34}.

\begin{figure}[htbp]%
\centering

\includegraphics[width=\columnwidth]{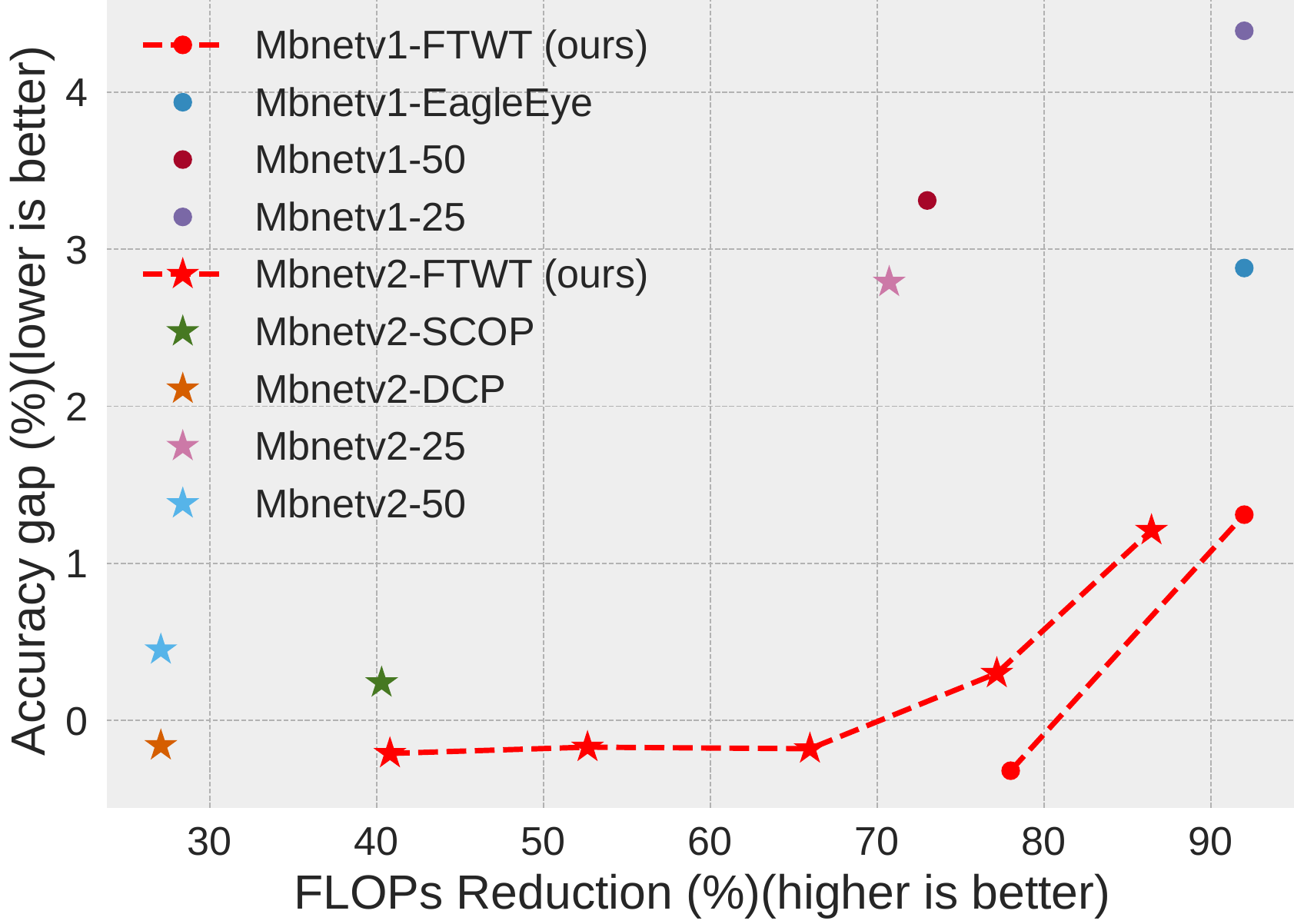}%
\caption{MobileNetV1/V2 on CIFAR10.}
\label{fig:resnet34}
\end{figure}

\subsection{Core Filters Visualization}
We show visualization of number of core filters per layer as explained in main papers. Core filters per layer indicate the filters that are activated in all different routes. Diversity is the highest at the middle layers which explains the accuracy drop in static uniform pruning (Table 1 main paper). Static MobileNet\_50 results on 3.31\% drop in accuracy in comparison to our dynamic method which improved the baseline with slight increase in accuracy 0.17\%.

\begin{figure}[htbp]%
\centering
\includegraphics[width=\columnwidth]{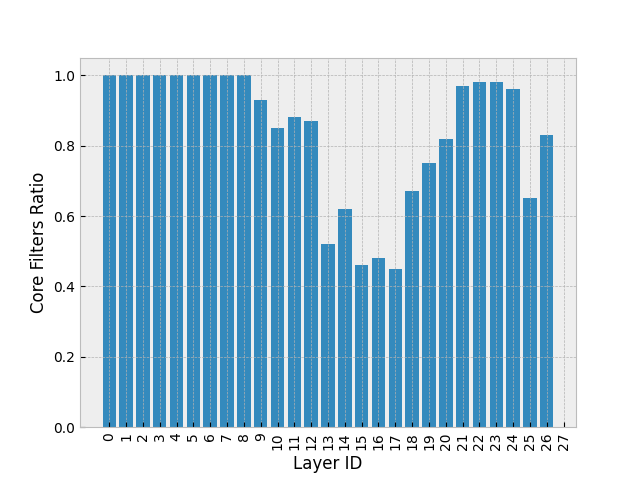}%
\caption{Number of core filters per layer in MobileNet.}%
\label{fig:corembnet}%
\end{figure}

\subsection{Error Bars}
Table \ref{tab:cifarruns} shows the numerical details of CIFAR experiments with the mean and standard deviation over 3 runs.

\begin{figure*}[!h]%
\centering
\includegraphics[width=2\columnwidth]{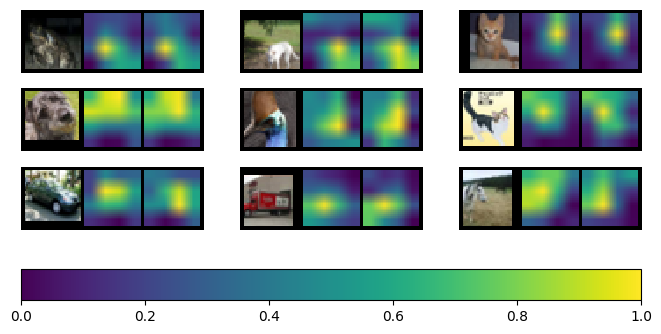}%
\caption{Heatmap visualization of random input samples from CIFAR for the 10th layer in MobileNetV1. Each triplet represents input image, baseline heatmap, pruned heatmap. FLOPs reduction in the layer is $\approx 70\%$, yet the pruned heatmap highly approximate the heatmap with fully activated      filters.}
\label{fig:heatmap}
\end{figure*}

\begin{table*}[]
\centering
\begin{tabular}{l|cc|cc|cc|cc}
\hline
Run & \multicolumn{2}{c|}{VGG16-BN} & \multicolumn{2}{c|}{VGG16-BN} & \multicolumn{2}{c|}{ResNet56} & \multicolumn{2}{c}{MobileNet} \\ \hline
 & Acc. (\%) & FLOPs (\%) &Acc. (\%) & FLOPs (\%) & \multicolumn{1}{l}{Acc. (\%)} & \multicolumn{1}{l|}{FLOPs (\%)} & \multicolumn{1}{l}{Acc. (\%)} & \multicolumn{1}{l}{FLOPs (\%)} \\ \hline
 & 93.26 & 73.10 & 93.66 & 65.41 & 92.61 & 66.45 & 91.01 & 78.03 \\
 & 93.21 & 73.00 & 93.49 &64.37 & 92.63 & 66.43 & 91.16 & 79.52 \\
 & 93.11 & 73.47 & 93.51 & 65.27& 92.65 & 66.41 & 91.03 & 78.04 \\ \hline
\textbf{Mean} & 93.19 & 73.19 &93.55 & 65.01&92.63 & 66.43 & 91.06 & 78.53 \\
\textbf{Std} & 0.07 & 0.24 &0.09 & 0.56& 0.02 & 0.02 & 0.08 & 0.8 \\ \hline
\end{tabular}

\caption{Mean and standard deviation across different runs on CIFAR.}
\label{tab:cifarruns}
\end{table*}

\subsection{Heatmap Visualization}
We visualize the heatmap of a highly pruned layer in comparison to the baseline model. Figure \ref{fig:heatmap} shows comparison between heatmap from the baseline with all filters activated and heatmap of dynamically selected filters. As can be seen, dynamic pruning approximates the baseline with high attention on foreground objects. This shows that even with 70\% pruning ratio in that layer, we are able to approximate the behavior of the original model.

\section{ImageNet}
\subsection{Joint vs Decoupled}
As training ImageNet models are expensive, we show few experiments to compare the results with joint and decoupled training modes in Table \ref{tab:imagenet}. Similar to results on CIFAR, decoupled training outperforms joint training under similar FLOPs reduction.

\begin{table}[!h]
\centering
\begin{tabular}{c|c|c|c}
\hline
\multicolumn{1}{l|}{Model} & Joint & Decouple & \multicolumn{1}{l}{FLOPs reduction (\%)} \\ \hline
\multirow{2}{*}{Resnet34} & 70.06 & 71.71 & 37 \\
 & 71.52 & 72.79 & 52 \\ \hline
\end{tabular}
\caption{Joint vs decoupled training on ResNet34 ImageNet}
\label{tab:imagenet}
\end{table}

\section*{Broader Impact}
Neural Network pruning has the potential to increase deployment efficiency in terms of energy and response time. However, obtaining these pruned models are yet to be optimized for a better overall computational consumption and more environment friendly. Moreover, pruning require careful understanding of deployment scenarios such as questioning out-of-distribution generalization \cite{liebenwein2021lost} or altering the behavior of networks in unfair ways \cite{paganini2020prune}. We showed results on out-of-distribution shift to tackle the first part. We did not investigate fairness of the model's prediction as both datasets (i.e CIFAR and ImageNet) are balanced. Although, we prune based on the mass of the heatmap equally for all samples. We hypothesize this trait can give our method an advantage over fixed pruning ratio which might hurt some input samples over some others.

\end{document}